\documentclass[conference]{IEEEtran}
\IEEEoverridecommandlockouts
\usepackage{cite}
\usepackage{amsmath,amssymb,amsfonts}
\usepackage{caption}
\usepackage{subcaption}
\usepackage{graphicx}
\usepackage{textcomp}
\usepackage{xcolor}
\usepackage{hyperref}
\usepackage[singlelinecheck=off]{caption}
\usepackage{algorithm}
\usepackage{algpseudocode}
\def\BibTeX{{\rm B\kern-.05em{\sc i\kern-.025em b}\kern-.08em
    T\kern-.1667em\lower.7ex\hbox{E}\kern-.125emX}}
\begin{document}

\title{ Optimal Video Compression using Pixel Shift Tracking}

\author{
    \IEEEauthorblockN{ Hitesh Saai Mananchery Panneerselvam*, Smit Anand*}
        \\    
    \IEEEauthorblockA{ hiteshsaaimananchery@gmail.com, smit.anand033@gmail.com }
}

\maketitle

\begin{abstract}
The Video comprises approximately ~85\% of all internet traffic, but video encoding/compression is being historically done with hard coded rules, which has worked well but only to a certain limit. We have seen a surge in video compression algorithms using ML-based models in the last few years and many of them have outperformed several legacy codecs. The models range from encoding video end to end using an ML approach or replacing some intermediate steps in legacy codecs using ML models to increase the efficiency of those steps.
    
Optimizing video storage is an essential aspect of video processing, so we are proposing one of the possible approaches to achieve it is by avoiding redundant data at each frame. In this paper, we want to introduce the approach of redundancies removal in subsequent frames for a given video as a main approach for video compression. We call this method Redundancy Removal using Shift (R\textsuperscript2S). This method can be utilized across various Machine Learning model algorithms, and make the compression more accessible and adaptable. In this study, we have utilized a computer vision-based pixel point tracking method to identify redundant pixels to encode video for optimal storage.
    
\end{abstract}


\section{Introduction}
Video compression algorithms are divided into two main categories, Lossy and LossLess. The lossy method is beneficial for transmission and storage since it provides a method to recover the compressed data with minimial data loss but the LossLess method is important for scientific and research data where data loss can have a huge impact on the data quality.

Training a single Neural Network to do the video compression as an end-to-end task has been proven to be quite a challenging task. Using a neural network to first convert a frame (or a bunch of frames) to a compressed blob and then try to recover the same input as an output ends up in missing or jumbled data. Research has been able to improve the working of Neural Networks in video compression tasks by using RNN, Residual Encoder, Variational Auto Encoder (VAE), and Deep Contextual Networks has shown improvements. There have been several attempts to use neural network models to replace fully or partially, the predictive or statistical models that have been historically used for video compression in several codecs and it has been able to improve the efficiency of the overall method
\cite{rippel2019learned, lu2019dvc, ma2019image, sullivan2005video, balle2016end}.

\textit{*Equal contribution. Listing order is random. The implementation of the code been done using single point tracking and it can be found \href{https://github.com/HiteshSaaiNew/optipixel_video_encode}{here}}

In this paper, we have tried to look at this problem differently by focusing on reducing the redundant pixel data that gets transferred from one frame to another frame in a video rather than compressing each image independently or an entire video in one go. This approach for compression has not been explored much. By using this approach, we have been able to propose a system that can be customized and used in many different algorithms, to have an optimal video storage system. The idea is to understand the similar pixels (or a collection of pixels) in subsequent frames and then label them as redundant. We will not store the redundant pixels but keep a memory of their shift positions for decompression. Another benefit of this approach is we just need a previous frame to decompress the next frame, which is good for video streaming since we don't need to process a lot of frames at the same time.

The way the compression works is, that we try to identify the movement of a pixel (or a collection of pixels) to understand their trajectory. By following their trajectory, we can estimate where the pixel is going to be in subsequent frames, then we try not to store the pixel (or collection of pixels) and just store its location in memory.

\section{Related Work}

\subsection{Using Recurrent Neural Network}
There has been a considerable amount of work related to Deep Learning-based video compression. RNN has been used extensively for this purpose because of its ability to hold context, \cite{toderici2015variable, toderici2017full} has used convolutional and de-convolutional LSTM recurrent networks where the encoder and decoder are stacked LSTM layers. \cite{kin2017video} have used a bidirectional RNN with GRU. They have used an Encoder-Decoder network to capture the statistical redundancies for lossless compression. \cite{yang2020learning} have used two Recurrent Auto Encoders (RAE) to get latent representation based on both current and previous input frames effectively decreasing the loss while decompression.

\subsection{Using Deep Convolutional Network (DConvN)}
\cite{yang2020learning, birman2020overview} have done a very good study explaining the different categories and how DConvN has been used to solve each category of video compression. \cite{yang2020learning} proposed Recurrent Learned Video Compression (RLVC) with Recurrent Auto Encoder (RAE) and Recurrent Probability Model (RPM) where the RAE employs recurrent cells in both the encoder and decoder. As such, the temporal information in a large range of frames can be used for generating latent representations and reconstructing compressed outputs. They have tried to use the conditional correlation between subsequent frames which reduces the entropy as compared to be calculated individually.  DeepStream was been proposed in \cite{amirpour2022deepstream} which is a scalable content-aware per-title encoding approach where bitrate-resolution pairs are upscaled to the resolution of the original video and the optimized bitrate-resolution pairs are selected to construct the bitrate ladder.

\subsection{Using Deep Contextual Network (DContN)}
This method has been used to move away from predictive coding to conditional coding \cite{li2021deep,guo2021causal,sheng2022temporal}. In these papers, authors are trying to either use temporal context or deep image context for encoding/reconstructing the frames of videos. In \cite{li2021deep} the authors, to tap the potential of conditional coding, have proposed using feature domain context as a condition. This enabled them to leverage the high-dimensional context to carry rich information to both the encoder and the decoder, which helps reconstruct the high-frequency contents for higher video quality.

\section{Proposed Method}

\subsection{Identifying Redundant Data using Pixel Tracking}

Redundancy removal using Shift (R\textsuperscript2S) works by calculating the shift of pixels (or a group of pixels) in subsequent frames. By calculating the shift, we can understand which pixels are repeated/redundant in the next frame and which are new. We will keep the new pixels and replace the repeated pixels with a black portion (ultimately reducing the storage size). \textit{We have used black pixels to replace the repeated/redundant pixels since we were storing the compressed images, one can also use any other data storage format like blob, or parquet to store the intermediate data which can further reduce the storage size}.

While decompressing, we can revert the shift and start retrieving the frames from the first frame. Decompression needs the first original/uncompressed frame with the shift data in memory. One good thing about the decompression procedure is, that we only need two frames at a time.We have used the pixel tracking approach as an implementation in this paper to show the effectiveness of our algorithm. Pixel tracking works by calculating the shift of the pixel movement from one frame to another. Then, based on the shift value of the pixel, we can identify the redundant and non-redundant (pixels that are also present in the first frame) pixels in the second frame, new pixels will be stored but older pixels can be removed. We will keep a memory shift of which block of pixels has been removed from each frame. 

We have an option to track a single pixel in videos where objects inside the video are not moving. This is a special case scenario but it's an easier way to showcase the viability of the approach. Otherwise, we can track multiple pixels in the video and create 2D grids in the frame, and for each given grid block we track the central pixel to calculate the shift and apply the shift to all pixels in the grid and follow the process for any given grid in the frame. There is also a costly option to track every pixel in the frame and not store pixels that didn't show a "significant" shift.

\subsection{Using Similarity Search Matrix}

A similarity search in a matrix has been used to retrieve the same types of images by querying. \cite{larsen2016autoencoding} has used a method that leverages learned representations to better measure similarities in data space. They used the combination of variational autoencoder (VAE) with a generative adversarial network (GAN) to use the learned feature representations in the GAN discriminator as a basis for the VAE reconstruction objective. \cite{chan2015progressive} used raw pixel to do the image similarity search between a database of images and a query image by finding the same group of pixels (which is in the query image) in the database image. They implemented the Nearest Neighbor (NN) search method to find a candidate \textit{c} of \textit{d} such that it has the smallest dist(\textit{q}, \textit{c}). Where \textit{q} is the query matrix and \textit{d} is the image in the dataset. \cite{gionis1999similarity} talked about hashing the images from the database to ensure that the probability of collision is much higher for objects that are close to each other than for those that are far apart.
\newline
\newline
This method can be used to find the similarity between two subsequent images by dividing the images into blocks and searching for similar blocks between the images. If a block from the next image has a similar block from the previous, then we don't need to store the block matrix in the next image and label it redundant, and eventually we reduce the storage space for a given video.

\subsection{Using Object Detection and Masking}
Another possible approach for optimizing video storage by avoiding redundant data at each frame is to utilize object detection using a deep convolution network. Based on various research studies on objects detection \cite{jiao2019survey, papageorgiou2000trainable, papageorgiou1998general, pathak2018application}, we were able to achieve detecting from various objects, humans, faces, etc. Based on the objects being detected we tend to get object pixel coordinates from an image, so we can leverage that to identify those repeated objects in every frame image of a video to reduce space, when it comes to objects we need to be specific on object shape, so coordinates alone won't be enough, so we might need to mask the object to track precisely, improve accuracy and reduce data loss for a optimal video decompression. The implementation of this approach would go through the Mask R-CNN method \cite{he2017mask}. We begin by briefly reviewing the Mask R-CNN detector. Mask R-CNN method consists of two stages, the first stage is called Region Propose Network (RPN) which proposes candidates and object bounding boxes, followed by the second stage in predicting the class and boxes offset and also parallelly outputs a binary mask for each candidate's bounding boxes. The subpart of candidate proposal and bounding box refinement prediction comes from the Faster R-CNN \cite{ren2015faster, dai2016r} method. So we can leverage this binary masking technique to identify redundant objects from one frame to the next frame sequentially and avoid storing those redundant Masked Object Pixels (MOP) from frame to frame. During the process of decompression, we would retrieve those missing MOPs (Masked Object Pixels) for the given frame from the previous frame's MOP to generate the completed current frame, by running it iteratively to retrieve missing MOP from the previous frame to complete the current frame, we would be able to decompress the video and also reduce the data loss on the decompression side.

To further enhance this approach we can just track the trajectories of the masked object which has moved from one frame to another. During the decompression phase, we just make use of the trajectory shifts to replace masked object pixels at each frame to decompress the video. This approach is more of an idea that utilizes the existing computer vision techniques to achieve optimal video storage with performance. By applying this method, we can reduce the storage space required for videos while maintaining their quality and performance.


\section{Implementation Detail}

\subsection{Pixel Point Tracking}
\subsubsection{\textbf{Single Point Trajectory}}
In this method, we rely on a specific technique for tracking points within a video sequence over an extended period. This technique involves selecting a reference point, denoted as coordinate \(\lambda = (x_{\text{1}},y_{\text{1}})\) from the initial frame as our pixel point for long-term single-point tracking to understand the camera motion shift for a given video. 

Our primary objective is to understand and quantify these camera motion shifts accurately. To achieve this, we employ a concept called point trajectory, which involves tracing the movement of the reference point over time. Using the point trajectory we optimize pixels per frame. We base our approach for point tracking using ”Persistent Independent Particles” (PIPs) \cite{ren2015faster} and PIPs++ \cite{zheng2023point}. Using PIPs we take X-fps RGB video as an input along with the target point coordinate to track. Using PIPs with sequel lengths as X we produce a \(1\text{x}2\) as an output coordinate for each frame. The output returns a shift (x,y) for a given point movement from the previous frame to the current frame. This process can be continued across each 8-frame sequence to obtain the coordinate shifts of the given tracking point across every frame in the video. \newline

\subsubsection{\textbf{Multi Point Trajectory}}
Single-coordinate point tracking is particularly effective in scenarios where the camera is in motion while the object of interest remains static. This method allows for precise tracking of a single point throughout the video, capturing the object's position relative to the moving camera. However, in situations where the object itself is in motion while the camera remains stationary, relying solely on a single coordinate point for tracking may not suffice.
When the object is in motion within the frame, a single tracking point may not adequately represent its movement, especially if the motion is complex or involves multiple directions. In such cases, using only one point for tracking could lead to missed or inaccurate detection of motion.

Consider a scenario where a camera is fixed in a room, capturing footage of people moving around. If the tracking algorithm focuses on just one point, it may fail to capture the full extent of the movement, resulting in incomplete tracking data.

To address this limitation, a multi-point tracking approach is necessary. By tracking multiple points of interest (POIs) simultaneously, the algorithm can capture a broader range of movement within each frame. This multi-point approach provides a more comprehensive understanding of the motion dynamics within the video, even when the camera remains stationary.

Incorporating Points of Interest (POIs) allows for the simultaneous tracking of multiple target points across the frame. These points can be strategically placed to cover different areas of interest within the frame, ensuring that no significant movement goes unnoticed.

By analyzing the trajectories of multiple POIs, the algorithm can determine the maximum shifting area within each frame accurately. This information is invaluable for optimizing pixel compression per frame, as it enables the algorithm to focus on the most relevant areas of motion, thereby reducing redundancy and optimizing storage efficiency. So we create 2D grid blocks for a given frame, and for each given grid block we track the central pixel to calculate the shift and apply the shift to all pixels in the given grid with movement and follow the process for any given grid in the frame, now we have a way to capture all the moving parts in a frame by divide and conquer method. There is also a costly option to track every pixel in the frame and not store pixels that didn't show a "significant" shift. 

\begin{figure}[!ht]
    \centering
    \captionsetup{format=hang, labelfont=bf}
    \includegraphics[width=1\linewidth]{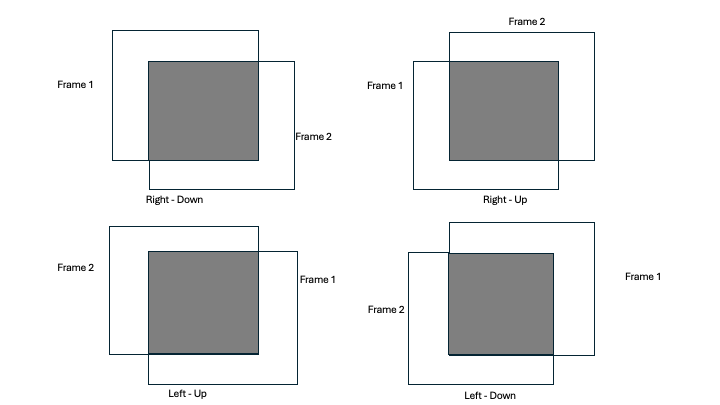}
    \caption{\textit{This picture shows the movement of frames/block. The grey area is the redundant part which doesn't need to be stored anymore}}
\end{figure}

\subsection{Compression Procedure}
In this initial research study, our focus was on optimizing video storage efficiency, particularly in scenarios where videos exhibit both dynamic camera movements and static elements within each frame. To achieve this, we utilized a single-point tracking technique as a proof of concept for redundancy approach. The process began by selecting an arbitrary tracking point coordinate from the initial frame. Subsequently, each frame of the video was systematically passed through a series of processing steps, utilizing PIPs (Picture-in-Picture) technology, we implemented the model predicting for every next frame to model predicting for a sequence of 8 frames in RGB video format, to validate the performance of tracking and data loss. \newline

Compression Per Frame,

\begin{equation} \label{eq1}
    \begin{split}
        \lambda_{\text{k}} = [x_{\text{n}},y_{\text{n}}]
    \end{split}
\end{equation}

\begin{equation} \label{eq2}
    \begin{split}
        \gamma(k) = (W-(W-x_{\text{n}})) + (H-(H-y_{\text{n}}))
    \end{split}
\end{equation}

Compression of total video, 
\begin{equation} \label{eq3}
    \begin{split}
        \zeta_{\text{compress}} \sum_{f=1}^{n} = \gamma(2) + \gamma(3) + \gamma(4) + .. + \gamma(n)
    \end{split}
\end{equation}

Where \(\lambda_{\text{k}}\) is the pixel point tracking shift in the kth frame, \(x_{\text{n}}\) and \(y_{\text{n}}\) is the x and y axis movement of the camera motion shift, \(\gamma(k)\) is the compression calculating function to store only the non-redundant pixels from the kth frame where as the W is the width and H is the height derived from the x,y shift movement and \(\zeta_{\text{compress}}\) is a total stack of all the compressed pixels from each frame. \newline

Through this process, we were able to derive the displacement in coordinate movements for each subsequent frame refer Fig.1. Take a look at Fig.2, for an illustration of compression per frame. Leveraging this displacement data, we were then able to discern between redundant and non-redundant dimensions of pixels within the frame. This allowed us to identify and isolate redundant pixel values. Remarkably, by nullifying these redundant pixels and assigning them a '0' pixel value, effectively representing black, we achieved a significant reduction in storage requirements per frame, amounting to a remarkable ~80-90\% decrease in storage space. Fig. 3(a) shows the performance of both \textit{pips \& pips++} with the data compression between 80-85\% and 87-95\% respectively and the Fig. 3(b) shows the data loss values for the same methods between 4-7\% for both \textit{pips \& pips++} .

Importantly, this compression process unfolded sequentially, frame by frame, ensuring a systematic and efficient reduction in storage demands across the entire video sequence. Ultimately, the compressed video, along with the associated tracking movement data of the selected coordinate point, was stored on disk, providing a streamlined yet comprehensive solution for optimized video storage in scenarios characterized by dynamic camera motion and static elements.

\subsection{Decompression Procedure}
In the context of decompression, the central aim is to reconstruct the original image or video frames from their compressed counterparts, which inevitably lose some information during the compression procedure. This loss typically manifests as missing pixels within the compressed frames. In our approach, compression begins from the second frame onward, leaving the initial frame uncompressed. This ensures that we have a reference point, free from any compression artifacts, for subsequent decompression. Consequently, after compression, the first frame is preserved in its entirety, while the following frames contain only the essential, non-redundant information, with the remainder of the frame appearing as blank or darkened.

\begin{figure}[ht]
    \centering
    \captionsetup{format=hang, labelfont=bf, justification=centering}
    \includegraphics[scale=0.25]{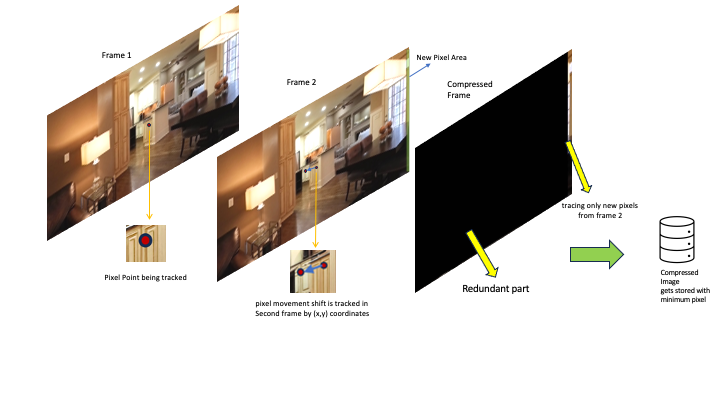}
    \caption{\textit{Representation of Frame compression using pixel point tracking}}
\end{figure}

Decompression Per Frame,
\begin{equation} \label{eq4}
    \begin{split}
        \hat{\lambda}_{\text{k}} = [x_{\text{n}},y_{\text{n}}]
    \end{split}
\end{equation}

\begin{equation} \label{eq5}
    \begin{split}
        PF = IF[CF_{\text{f}}=CF_{\text{2}},OF_{\text{f-1}},DF_{\text{f-1}} ]
    \end{split}
\end{equation}

\begin{equation} \label{eq6}
    \begin{split}
        \psi_{\text{f-1}} = PF\{x_{\text{n}},W\} +  PF\{y_{\text{n}},H\}
    \end{split}
\end{equation}

\begin{equation} \label{eq7}
    \begin{split}
        \Gamma(k) = \psi_{\text{f-1}} + \xi_{\text{f}}
    \end{split}
\end{equation}

Decompression of total video, 
\begin{equation} \label{eq8}
    \begin{split}
        \Psi_{\text{decompress}} \sum_{f=1}^{n} = \Gamma(1)+\Gamma(2)+ ...+ \Gamma(n)
    \end{split}
\end{equation}

Where \(\hat{\lambda}_{\text{k}}\) is the inverse of the pixel point tracking shift (4), PF, CF, OF is the previous frame, current frame, and original frame (5), \(\psi_{\text{f-1}}\) data retrieval calculation for frame f-1, \(\Gamma(k)\) is the decompression of frame using previous frame and current non-redundant frame and finally \(\Psi_{\text{decompress}} \sum_{f=1}^{n}\) is the summation of all decompressed frame data compiled back together to retrieve the video.\newline

\begin{figure*}[ht]
     \centering
     \captionsetup{format=hang, labelfont=bf}
     \begin{subfigure}{0.45\textwidth}
         \centering
         \captionsetup{justification=centering}
         \includegraphics[width=1\linewidth]{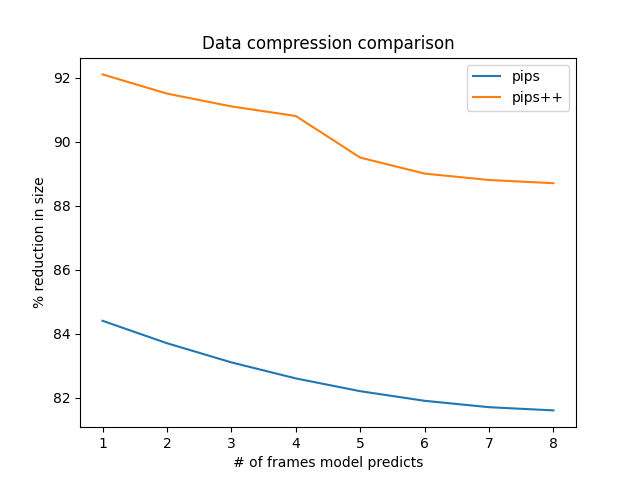}
         \caption{Data size Compression Percentage}
     \end{subfigure}
     \hspace*{\fill}
     \begin{subfigure}{0.45\textwidth}
         \centering
          \captionsetup{justification=centering}
         \includegraphics[width=1\linewidth]{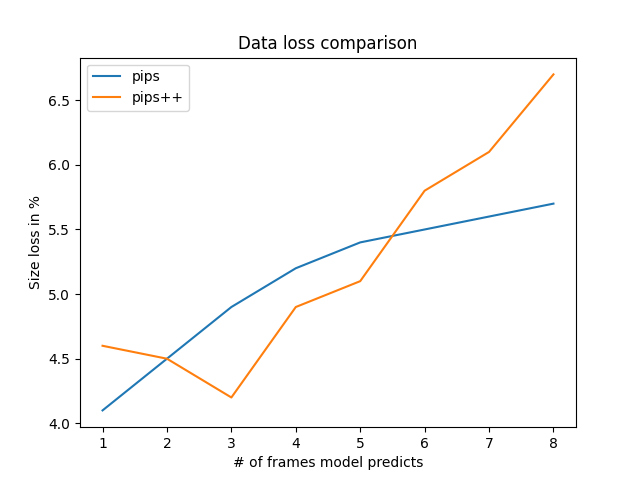}
         \caption{Data loss percentage}
     \end{subfigure}
     \hspace*{\fill}
        \caption{{\textit{Picture (a) shows the compression percentage (size of compressed image  w.r.t the original size of image) w.r.t the number of frames for which the model is predicting the shift. Picture (b) shows the data loss w.r.t the number of frames for which the model is predicting the shift. We see a big difference in the size reduction after compression between pips and pips++ but we that data loss in pip++ was high initially, it went down for few frames and then it went up at the end compared to pips. Its because, in our experiment, pips++ performed better when predicting the shift in the pixels. These values will vary based on the model and video of choice and point of tracking method. This graph is created by running pips model with single point of tracking on a 1920*1080 video using 30 frames at 30fps}}}
        \label{fig:three graphs}
\end{figure*}


To facilitate the reconstruction process, we rely on two key components: the compressed video data and the stored records of point track movement. These movement records provide valuable information about how each point in the video frame has shifted over time, allowing us to understand the dynamics of motion within the scene. During decompression, for each compressed frame N, we retrieve the corresponding movement shift data \( \lambda = [x_{\text{n}},y_{\text{n}}]\) from the stored records. These shift values indicate the direction and magnitude of movement for each pixel in the frame. Importantly, these shifts are reversed during decompression to reconstruct the original pixel positions. \newline

\begin{figure}[ht]
    \centering
    \captionsetup{format=hang, labelfont=bf}
    \includegraphics[scale=0.25, trim={3cm 4cm 0 0}]{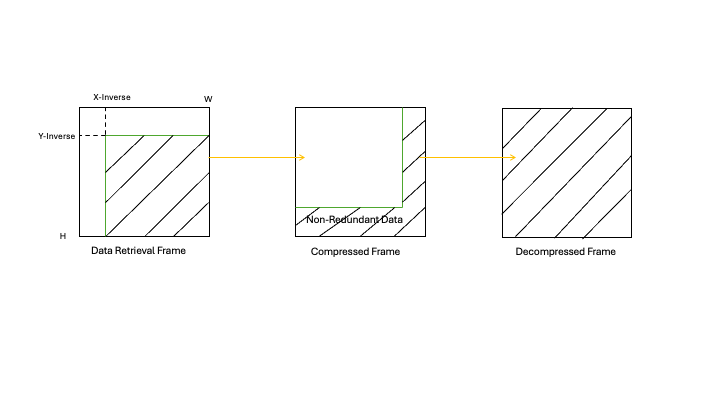}
    \caption{\textit{Data Retrieval Procedure for decompression per frame}}
\end{figure}

Using the inverse of the shift values \( \hat{\lambda}\), we determine the boundaries (height and width) from which to extract pixel values in the preceding frame. By referencing the previous frame, we effectively 'fill in' the missing pixels in the current frame based on their original positions. This process ensures that the decompressed frame retains as much detail and fidelity as possible.Once the missing pixels have been restored, the decompressed frame is reconstructed. This iterative process is repeated for each frame until all frames are decompressed, ultimately culminating in the reconstruction of the video sequence. The result is a reconstruction of the original video, free from the compression artifacts that were introduced during the compression process. To sum it up, Fig.4 shows the frame/block of frame movement and how we can calculate the redundant data for compression and later retrieve that data for decompression.

\begin{algorithm}
\caption{Pseudo code for both compression and decompression by R\textsuperscript2S using Pips}\label{alg:cap}
\begin{algorithmic}
\Require $S \geq 2$ Stride length
\Require $N \geq 1$ Number of points to track
\Require $M \doteq (N of frames) // (S // 2) - 1$ Max iteration, Overlapping max iterations to run
\State $G \gets 0$ global step
\While{$G \leq M$}
    \State Read and append the Stride number of image in a matrix
    \State Call model for trajectory prediction
    \State $C \doteq 0$
    \While{$C \leq S$}
    \State Call compression with shift from the model trajectory \eqref{eq2}
    \State Save the compressed matrix
    \State Save the shift in memory
    \State Call Decompression and pass the shift from the memory \eqref{eq5}
    \State Save the decompressed image
    \State $C = C + 1$
    \EndWhile
\EndWhile
\end{algorithmic}
\end{algorithm}

\section{Experiments \& Results}
We have experimented with single-point tracking on a video using the Pips and Pips++ approach. We chose the single-frame prediction approach, which means, we are asking the model to predict a single point, for example (100, 450) pixel coordinate for x number of frames. This will be the slowest method since the model is getting called for each frame, but the most accurate one also. You can ask the model to predict the trajectory for the next `n` frames and then skip those frames from the next iteration, this will make the prediction a little less accurate but faster.

\begin{table}[ht]
    \centering
    \captionsetup{justification=centering}
    \begin{center}
        \begin{tabular}{|c|c|c|}
            \hline
            \textbf{Method} & \multicolumn{2}{|c|}{\textbf{Speed}} \\
            \cline{2-3}
            \textbf{}  & \textbf{\textit{tpf$^{\mathrm{a}}$}}           & \textbf{\textit{size in KB}}  \\
            \hline
            Compression           & 15                   &           36.82               \\
            \hline
            Decompression           & 50                   &          238.57                  \\
            \hline
            \multicolumn{3}{l}{$^{\mathrm{a}}$time per frame in ms}
        \end{tabular}
    \caption{\textit{Comparison on optimal video storage for different fps video}}
    \end{center}
\end{table}

The following table I contains the time per frame that it takes for the compression and decompression to work with single-point tracking. The experiment has been performed on a system with a 2.6 GHz intel core i7 with 32 GB 2667 MHz RAM.

In Fig. 3(a), we can see that the compression performance deteriorate when we increase the Stride (the number of frames for which the model is predicting the trajectory). This is because, every new frame added increases the error values in prediction with the next frame which reduces the shift accuracy which results in averaging of incorrect shift. When the shift gets deviated far away from actual shift, which can lead to more compressed data but the result could be more data loss after decompression, refer to Fig. 3(b). From our experiment and observation, higher the accuracy with less data loss comes with smaller stride value.

Add for total video size comparison frame size 1080*1920 for 30 fps. The initial size of the uncompressed frame is 250.67 KB. There is few loss of data we encountered which can be attributed to the coordinate shift coming from the model. Since this is a single-point tracking data, the loss is expected to be more, but it can always be reduced by changing from single-point tracking to multi-point tracking using 2D grid method and the loss can be reduced significantly.

\bibliographystyle{plain}
\bibliography{reference}

\vspace{12pt}
\color{red}
\end{document}